\documentclass{article}
\usepackage{PRIMEarxiv}
\usepackage[utf8]{inputenc} 
\usepackage[T1]{fontenc}    
\usepackage{hyperref}       
\usepackage{url}            
\usepackage{booktabs}       
\usepackage{amsfonts}       
\usepackage{nicefrac}       
\usepackage{commath}
\usepackage{enumitem}
\usepackage{caption}
\usepackage{xcolor}
\usepackage{subcaption}
\usepackage{microtype}      
\usepackage{fancyhdr}       
\usepackage{graphicx}       

\title{Denoised Labels for Financial Time-Series Data via Self-Supervised Learning}

\author{
  Yanqing Ma \\
  King's College London \\
  London, United Kingdom\\
  \texttt{yanqing.ma@kcl.ac.uk} \\
   \And
  Carmine Ventre \\
  King's College London \\
  London, United Kingdom\\
  \texttt{carmine.ventre@kcl.ac.uk} \\
   \And
  Maria Polukarov \\
  King's College London \\
  London, United Kingdom\\
  \texttt{maria.polukarov@kcl.ac.uk} \\
}

\begin{document}
\maketitle

\begin{abstract}
The introduction of electronic trading platforms effectively changed the organisation of traditional systemic trading from quote-driven markets into order-driven markets. Its convenience led to an exponentially increasing amount of financial data, which is however hard to use for the prediction of future prices, due to the low signal-to-noise ratio and the non-stationarity of financial time-series. Simpler classification tasks --- where the goal is to predict the directions of future price movement --- via supervised learning algorithms, need sufficiently reliable labels to generalise well. 
Labelling financial data is however less well defined than other domains: did the price go up because of noise or because of signal? The existing labelling methods have limited countermeasures against noise and limited effects in improving learning algorithms. 
This work takes inspiration from the image classification in trading \cite{image1} and success in self-supervised learning (e.g., \cite{self1}). We investigate the idea of applying computer vision techniques to financial time-series to reduce the noise exposure and hence generate correct labels. We look at the label generation as the pretext task of a self-supervised learning approach and compare the naive (and noisy) labels, commonly used in the literature, with the labels generated by a denoising autoencoder for the same downstream classification task. Our results show that our denoised labels improve the performances of the downstream learning algorithm, for both small and large datasets. We further show that the signals we obtain can be used to effectively trade with binary strategies \cite{murphy1999technical}. We suggest that with proposed techniques, self-supervised learning constitutes a powerful framework for generating "better" financial labels that are useful for studying the underlying patterns of the market.
\end{abstract}


\section{Introduction}
The notion of economic noise was introduced in 1986 by Fischer Black \cite{noise}, who describes the term as information with opposite effects on pricing, such as hype and inaccurate data. In financial markets, noise is considered as a form of market movement and can be the result of unknown interactions between the many actors within (e.g., algorithmic trading, large investment institutions). Noise can not only distort the market trend and increase the possibility of trading on false signals, but also damage an investor's long-term portfolio performances due to possible over-reaction. Once the underlying trend is identified, the future price movement can be predicted more efficiently. Conventional strategies to reduce noise exposure include applying different technical indicators, including, for example moving averages, before making trading decisions. 

More recent approaches to predict future price movement are concentrating on using supervised learning algorithms, where labels are required for training the model. Labels are defined based on certain properties such that the useful features of the financial time-series can be learned and then generalised on unseen samples. Labelling is not too controversial in other domains (e.g., a picture either contains or not a cat) and, in fact, approaches in the computer vision area are well defined and extensively studied, including semantic segmentation (e.g., \cite{DBLP:journals/corr/ShelhamerLD16}) and panoptic segmentation (e.g., \cite{DBLP:journals/corr/abs-1801-00868}). On the contrary, financial time-series is hard to label universally and effectively. The current labelling methods for financial data can be divided into two groups used for static or dynamic markets, depending on whether or not they include volatility. For example, the binary labelling method and the fixed-time horizon method assume that the market remains static, whereas the triple-barrier method and quantised-labelling method are used to process the dynamic changes in the market \cite{machine_learning_book}.

Self-supervised learning has been widely adopted in natural language processing (e.g., \cite{self1,DBLP:journals/corr/abs-1810-04805,DBLP:journals/corr/abs-1811-06964}) and computer vision (e.g., \cite{DBLP:journals/corr/abs-1901-09005,DBLP:journals/corr/abs-1804-03641}). The algorithm is proposed to extract useful features from large-scale unlabelled datasets so that labels can be generated without human annotation. The self-supervised learning process is divided into two successive tasks, namely, pretext task and downstream task, and the model works by training the unlabelled dataset through pretext tasks and then transferring the learned parameters into downstream tasks. In computer vision, pretext tasks are some unsolved pre-designed tasks so that useful representations can be learned by solving the tasks. Downstream tasks are algorithms that can examine the quality of representations learned during pretext task \cite{self1}. A denoising autoencoder is a paradigm adopted for pretext tasks to reconstruct corrupted images (e.g., \cite{denosing_example}). Other popular methods used for image recognition are image colourisation (e.g., \cite{DBLP:journals/corr/ZhangIE16}), context encoder (e.g., \cite{DBLP:journals/corr/PathakKDDE16}) and contrastive learning (e.g., \cite{DBLP:journals/corr/abs-1807-03748}). 

After seeing the efficient bottleneck network structure of denoising autoencoder in image recognition to automatically generate labels, it is natural to ask whether self-supervised learning can be applied to produce denoised labels for financial time series. Given the non-convergent nature of financial data, one key question is how to interpret the results. What does it really mean for the labels to remove (or less ambitiously reduce) noise? In our study, we will make this question more precise, as follows:
\begin{enumerate}[topsep=5pt, itemsep=5pt]
    \item Will labels make prediction easier?
    \item Can we reliably take positions on these predictions? 
\end{enumerate}
To answer these questions, we design a contrastive experiment evaluating the performance of having denoising autoencoder as a pretext task versus naive labels for a classification downstream task, for which we use a simple Support Vector Machine (SVM).

\section{Main Contributions}
To the best of our knowledge, we are the first to apply principles of self-supervised learning to financial time series. Technically speaking, our contrastive experiments test the advantages of considering the raw financial time-series data as corrupted inputs for an autoencoder, that outputs reconstructed time series. Our results allow us to give answers to the two questions above.
\begin{enumerate}[topsep=5pt, itemsep=5pt]
    \item The labels generated by the autoencoder generally make prediction of future price movement easier for the SVM classifying downstream, with an increase of 20\% or more in terms of F1 score, depending on the classes we adopt. 
    \item The labels generated by the autoencoder generally preserve the trading signals of a class of binary trading strategies \cite{murphy1999technical}. They seem to avoid taking unnecessary positions, and make execution cheaper.
\end{enumerate}
The benchmark we use in our experiments is a naive labeling method, whereby market prices are taken at face value and used to define the classes up/stationary/down. Despite its simplicity, this method is widely used in the literature and by practitioners. Interestingly, the validity of the output of the autoencoder can also be assessed visually by comparing the original financial time series with the reconstructed time series, the latter being smoother and less bumpy. 

The observations at point (1) above are also robust to different latencies and asset classes, as we prove them valid on different datasets (see below for more details). 

Ultimately, our experiments indicate that self-supervised learning can be used to produce high quality denoised  labels that can be used to improve the use of machine learning techniques in finance.

\section{Related Work} 
This work aims at analysing financial time-series data with a widely used algorithm from computer vision. Classical approaches of classifying financial time-series adopt the principles from physics perspective and focus on mathematical derivation. For example, wavelets (e.g., \cite{doi:10.1142/S0218194016400088,10.1007/978-3-540-24775-3_71}) and Fourier transform (e.g., \cite{StatisticalMethods}) are used to identify the signals in the frequency domain. Distance-based methods coupled with similarity metrics are another popular approach in time-series classification, such as the nearest neighbour classifier with the dynamic time warping distance function \cite{cite-key}. In \cite{deeplearningreview}, the authors present the empirical study of the performance of different deep neural networks for classifying the univariate and multivariate time-series dataset. In addition to this,  deep learning algorithms, such as recurrent neural networks with Long Short-Term Memory (LSTM) units \cite{doi:10.1080/14697688.2019.1622295, 8081663} and stacked autoencoders with LSTM units \cite{10.1371/journal.pone.0180944}, are adopted for predicting the future price movement from large-scale high-frequency datasets.

Recent research suggests to solve the time-series classification problem by encoding the time-series data as a form of images and then adopting the techniques from computer vision (e.g., \cite{Park_2019,Wang2014EncodingTS,DBLP:journals/corr/WangO15}). There are examples of financial time-series, where the candlestick charts were encoded into images to replicate the human's trading strategies \cite{image1} or to identify the candlestick patterns \cite{DBLP:journals/corr/abs-1901-05237}. Another work focused on transforming financial ratio data into images before feeding them into the convolutional neural network \cite{9222178}. Furthermore, \cite{DBLP:journals/corr/abs-2002-09545} proposed to detect the anomalies in time-series by making use of an encoder-decoder architecture with skip connections.

\section{Data and Methods}
We start with the daily close price of S\&P 500 index from Yahoo finance for the period 2017-2019 to represent the case with a small dataset available (i.e., 502 observations in total). We then use Bitcoin 1-minute price \cite{bitcoin} for the period from Jun 2020 to Jun 2021, and high-frequency limit order data of Google on the day Jun 21st 2012 \cite{hflob} to represent the case with a large and noisy datasets. 

We let $X$ be the feature matrix of $n$ daily close prices $x(t)$ of the given dataset with $t=1,...,n$. The simple return of this asset will be $$R_{t+1}=\frac{x({t+1})-x({t})}{x({t})}.$$ To analyse the relative variations between two successive business days, the log return \cite{StatisticalMethods} is defined as following: 
\begin{align*}
r(t+1) & = \log\left(\frac{x(t+1)}{x(t)}\right) \\ &= \log[x(t+1)] -\log[x(t)].    
\end{align*}
The observation $x(t)$ is then assigned a label $y_{t} \in \{-1,0,1\}$ \cite{machine_learning_book}, which is computed by:
\begin{equation}\label{eq:naive}
    y_{t} = \begin{cases}
1 & $if$\  r(t)>\tau\\
-1 & $if$\ r(t)<-\tau\\
0 & $if$\ \abs{r(t)} \leq \tau \end{cases}
\end{equation} 
where $\tau$ is a threshold value between $(\min(r),\max(r))$. When $\tau$ equals to zero, the setup becomes that of a binary classification problem with labels \{-1,1\} only. We call the labeling method in \eqref{eq:naive} \emph{naive}; it is a special case of  fixed-horizon methods that is often used in practice \cite{machine_learning_book}.

The contrastive experiment is set up with two different workflows as described below.

\medskip \noindent {\bf Workflow 1.} The first workflow benchmarks the performance of financial time-series classification by using SVM only. The default values of parameters $C$ and "gamma" are used for SVM with the kernel of Radial Basis Function (RBF). The model is trained on the first 80\% of observations ${X}$ with corresponding labels and tested on the last 20\% of the dataset.

\begin{figure*}[ht]
  \centering
  \includegraphics[width=0.9\linewidth]{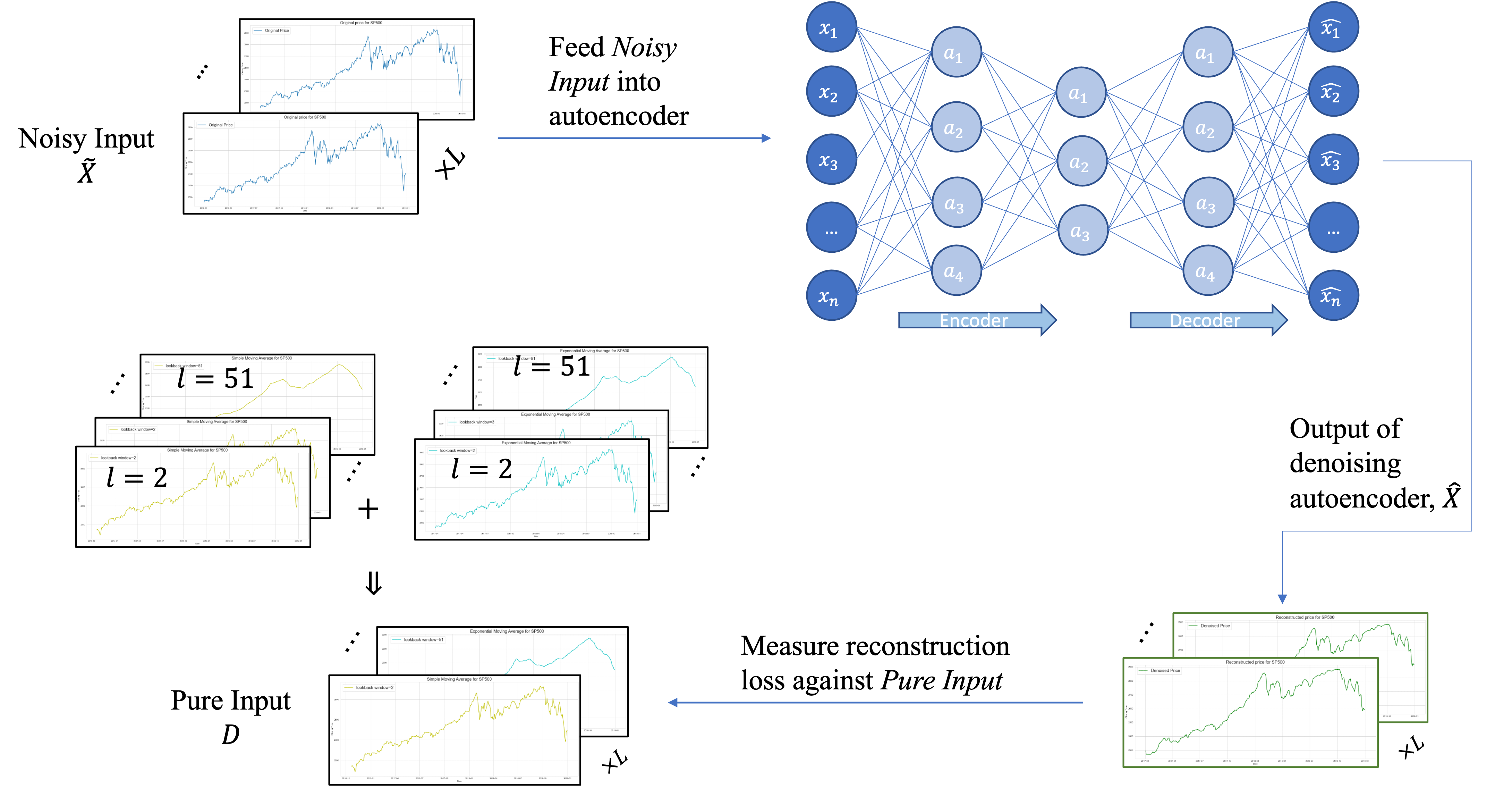}
  \caption{The Framework of Pretext Task of Workflow 2}
  \label{fig:pretexttask}
\end{figure*}

\medskip \noindent {\bf Workflow 2.} The second workflow implements self-supervised learning on real financial time-series data by using a denoising autoencoder for the pretext task and SVM with the same hyperparameters as Workflow 1 for the downstream task. We expand the feature matrix ${X}$ from a 1-dimensional column array to a
multi-dimensional matrix, so that the datasets are ready for the denoising autoencoders without overfitting \cite{Goodfellow-et-al-2016}. The denoising autoencoder is designed to have two types of input, the original input and the corrupted input, which are then trained through the 1-d convolutional neural networks. To avoid confusion, we named the new feature matrix as ``pure input'', $D$, and corrupted input as "noisy input", ${\Tilde{X}}$. The pure input is composed of moving averages with different lookback windows as follows, 
\begin{equation} \label{eq:pure_input}
    {D} = 
    \begin{bmatrix}
    SMA_{1}^{l_{2}}  & ... & SMA_{1} ^{l_{k}} & EMA_{1} ^{l_{2}} &  ... & EMA_{1} ^{l_{k}}\\
     SMA_{2} ^{l_{2}} \rule{0ex}{3ex} & ... & SMA_{2} ^{l_{k}} & EMA_{2} ^{l_{2}} & ... & EMA_{2} ^{l_{k}}\\
     ... & ... & ... & ... & ... & ...\\
    SMA_{n} ^{l_{2}} \rule{0ex}{3ex} & ... & SMA_{n} ^{l_{k}} & EMA_{n} ^{l_{2}} &  ... & EMA_{n} ^{l_{k}}\\
\end{bmatrix}^T.
\end{equation}
$D$ is a ${L} \times n$ matrix, where $l_i \in (2,n]$ is the length of lookback window for $i=2,...,k$, and ${L}=2 (l_k-l_2 +1)$ is the total number of different moving averages. We use moving averages because of their higher degrees of smoothing. The pure input matrix can be structured with simple moving averages (SMA) only, exponential moving averages (EMA) only or a combination of both. Based on our experiments, the combined structure performs better than the other two options. The corrupted input, ${\Tilde{X}}$, is composed of $L$ identical daily close price, which is due to the fact that financial time-series data is noisy enough and noise signals are random. It is built up as follows:
\begin{equation} \label{eq:noisy_input}
    {\Tilde{X}}=
    \begin{bmatrix}
    x_1 & x_1 & ... & x_1 \\
    x_2 & x_2 & ... & x_2 \\
    ... & ... & ... & ...\\
    x_n & x_n & ... & x_n \\
    \end{bmatrix}^T, \; {\Tilde{X}} \in {L}\times n \times {1}
\end{equation}

\begin{figure*}[ht]
  \centering
  \includegraphics[width=0.9\linewidth]{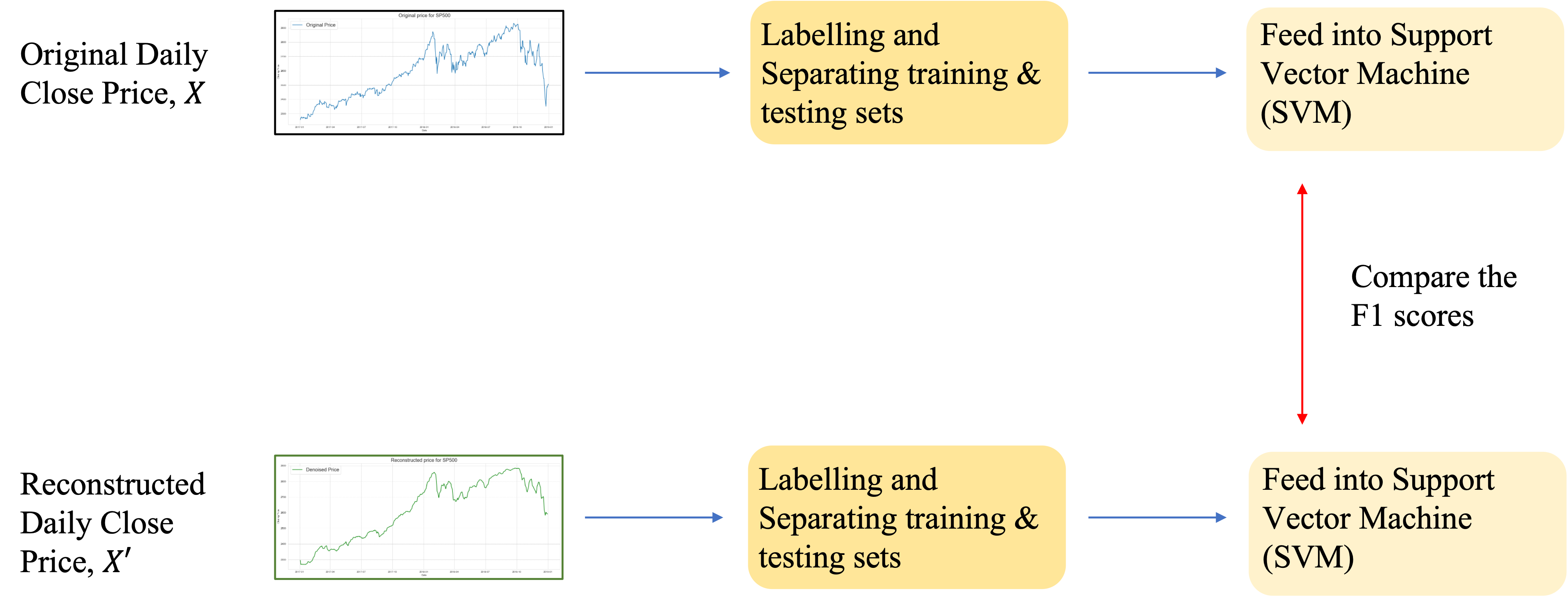}
  \caption{The Framework of Downstream Task of Workflow 2 (bottom) and the Framework of Workflow 1 (top)}
  \label{fig:downstreamtask}
\end{figure*}

Figure \ref{fig:pretexttask} describes the framework of pretext task, where the noisy input ${\Tilde{X}}$ in (\ref{eq:noisy_input}) is fed into the denoising autoencoder. The neural network is composed of two Conv1D layers as an encoder, two Conv1DTranspose layers as a decoder, and one Conv1D layer with the output, the Sigmoid activation function and the padding as an output layer. It is trained by minimising the reconstruction loss, which measures the difference between the pure input (\ref{eq:pure_input}) and output $\hat{X}$. The output matrix, $\hat{X}$, has the same dimension as pure input (\ref{eq:pure_input}) and it can be expressed as follows:
\begin{equation} \label{eq:output}
     {\hat{X}}=
    \begin{bmatrix}
    \hat{x_1} & \hat{x_1} & ... & \hat{x_1} \\
    \hat{x_2} & \hat{x_2} & ... & \hat{x_2} \\
    ... & ... & ... & ...\\
    \hat{x_n} & \hat{x_n} & ... & \hat{x_n} \\
    \end{bmatrix}^T
\end{equation}
Since the output $\hat{X}$ represents the denoised outcome of noisy input ${\Tilde{X}}$ and the noisy input is built up with $L$ identical daily close price, we break down the output $\hat{X}$ as $L$ identical column matrices ${X'}$ of $n$ denoised daily close price $\hat{x}(t)$ with $t=1, ..., n$. The bottom part of Figure \ref{fig:downstreamtask} illustrates the framework of downstream task. The training and testing labels are both updated during the downstream task since the labels are computed based on the reconstructed daily close price, ${X'}$, and the same parameters are used in SVM.


We examine the performance of the denoised daily close price by applying three commonly used trading indicators and we compare the triggered buying signals only between the denoised daily close price and the original daily close price. The buying signals are defined as the following:
\begin{itemize}[topsep=10pt,itemsep=5pt]
\item Moving Average crossover: Moving Average (MA) Crossover is used to indicate the change in trend and trading opportunities where the shorter simple moving averages cross the longer simple moving averages. It has greater degrees of lag comparing to single moving averages but reduces the loss of changing in directions. The indicator is composited of two simple moving averages with a shorter period and a longer period, respectively. For example, the shorter moving average could have 5-,10-, or 25-day period while the longer moving average commonly have 50-,100- or 200-day period. A buying signal is generated when the shorter moving average crosses the longer moving average as it suggests upward momentum \cite{murphy1999technical}. 

\item Moving average convergence divergence: Moving average convergence divergence (MACD) is similar to MA crossover, which is used as a trend following indicator by measuring the difference between two exponential moving averages. The indicator consists of a faster line and a slower line, named MACD line and signal line correspondingly. The faster line (i.e., the MACD line) measures the price velocity \cite{Encyclopedia} by subtracting the 26-day exponential moving average away from the 12-day exponential moving average. The MACD is positive when the 12-day moving average is above  the 26-day one and it indicates upward momentum. On the contrary,  the negative MACD represents downward momentum when the 26-day moving average is above the 12-day one. The slower line (i.e., the signal line) is used intentionally to smooth the MACD line further, and the 9-day exponential moving average of the MACD line is commonly used by analysts. A buying opportunity is triggered when the MACD line crosses above the signal line \cite{murphy1999technical}.

\item Bollinger Bands: Bollinger Bands (BB) is a technical indicator widely used for momentum trending following and detecting overbought and oversold conditions. It can describe the price volatility over time by including a graphical band on a two-dimensional graph. The indicator consists of an upper band, a middle band and a lower band, where upper and lower bands are resistance line and support line of the financial time-series, respectively \cite{Encyclopedia}. The middle band is normally the 20-day simple moving average and it is enveloped by the upper and lower bands. The upper band is twice a 20-days standard deviation above the middle band and conversely, the lower band is twice a 20-day standard deviation below the middle band. A buying signal is generated when the price line crosses above the lower band since it indicates a potentially oversold situation \cite{murphy1999technical}.
\end{itemize}

\section{Results}
This work intends to analyse whether or not self-supervised learning is useful for financial time-series classification so that noise signals can be removed efficiently. We conduct empirical studies to verify the effectiveness of the denoised close price in predicting the future price movement and to investigate whether or not the true signals are smoothed out at the same time.

\subsection{Small low-frequency dataset}
We begin by discussing the results for the S\&P 500 dataset.

\begin{figure*}[!h]
  \centering
  \includegraphics[width=0.8\linewidth]{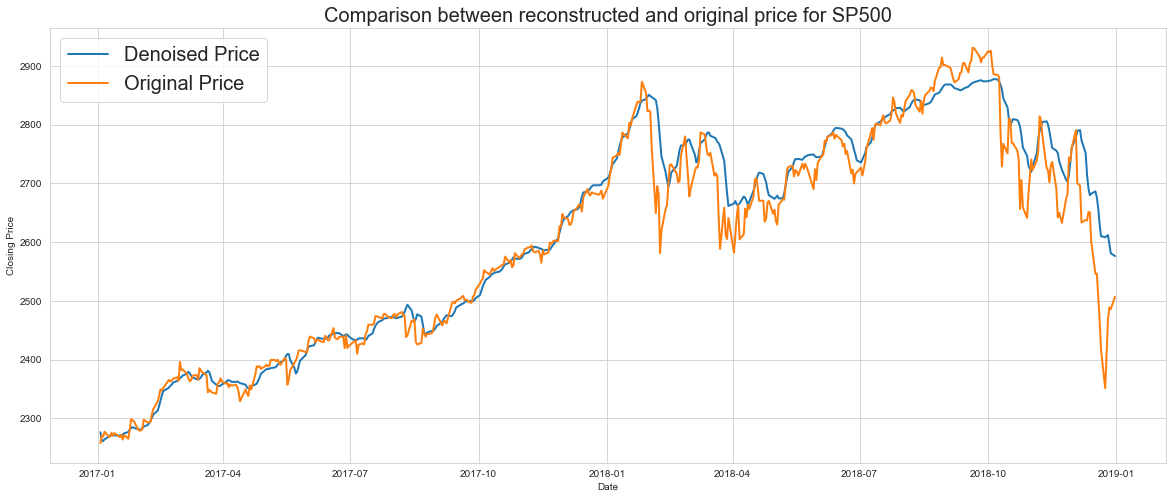}
  \caption{The Comparison Between Reconstructed and Original Daily Close Price of S\&P 500}
  \label{fig:smoothedcurve}
\end{figure*}

\smallskip \noindent {\bf Visual results.} Figure \ref{fig:smoothedcurve} describes the comparison between the original S\&P 500 daily close price (i.e., the orange curve) and the denoised price (i.e., the blue curve). It is clear to see that the denoised price is smoother than the original one, which shows a positive sign that the denoising autoencoder can recognise the noise signals from the corrupted input. In addition, the denoised price curve has shorter lags compared to simple moving averages, but higher degrees of smoothing compared to exponential moving averages. 
This graph supports our conclusion that the autoencoder in Workflow 2 goes beyond conventional approaches to denoise the data.  

We next take a more quantitative approach in assessing the quality of the reconstructed time series.

\begin{figure*}[ht]
    \centering
   \includegraphics[width=0.8\linewidth]{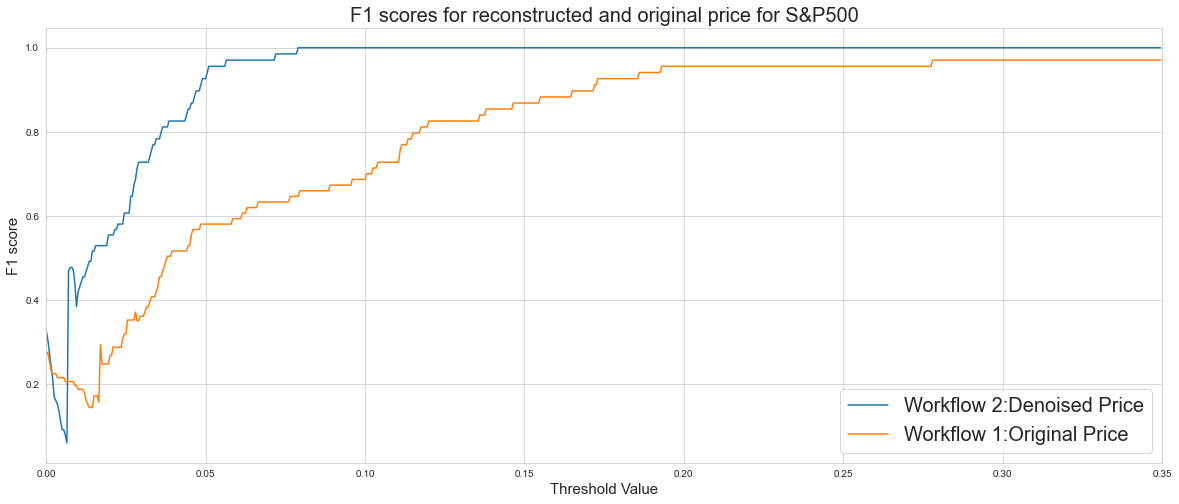}
   \caption{The comparison of F1 scores of the reconstructed daily close price and original daily close price}
   \label{fig:Ng10}
\end{figure*}

\begin{figure*}[ht]
    \centering
    \includegraphics[width=0.8\linewidth]{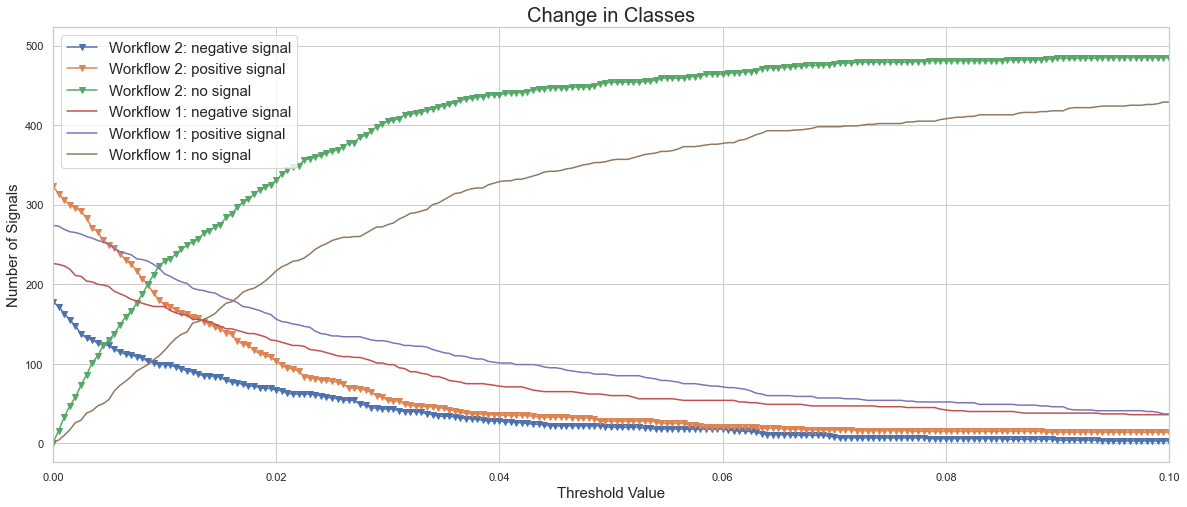}
    \caption{The change of number of signals per class with respect to threshold value}
    \label{fig:Ng9}
\end{figure*}

\smallskip \noindent {\bf Effect on downstream task.}
 Figure \ref{fig:Ng10} illustrates the F1 scores of both workflows. Under the same threshold $\tau$, Workflow 2 achieves higher F1 score and growth with faster speed than Workflow 1. This result is explained by Figure \ref{fig:Ng9}: there are more samples in Workflow 2 being classified into the ``no signal'' group at the same threshold $\tau$ level. This indicates that many labels are misclassified into either of the other two groups under the same threshold $\tau$ due to the noise and the denoising autoencoder successfully recognised and removed the noise. Therefore, the prediction of the future price movement has higher accuracy scores and performs better. It is worth mentioning that we label the prices using the fixed-time horizon method \cite{machine_learning_book} as the performance of this method is negatively affected by the noise and so is gradually replaced with volatility-based methods. Especially for the returns between two consecutive days, the fixed-time horizon method may cause confusion when analysing the underlying trend and hence, the buying and selling positions changes frequently and costly. For example in Table \ref{tab:bb}, the BB indicator suggests buying on 2018-02-05 and 2018-02-08 based on the original close price and buying on 2018-02-08 only based on the denoised close price. Importantly, this proves that the denoised close price can also be beneficial for traders in decision-making by reducing the number of false signals.

\begin{figure*}[ht]
\centering
\includegraphics[width=0.8\linewidth]{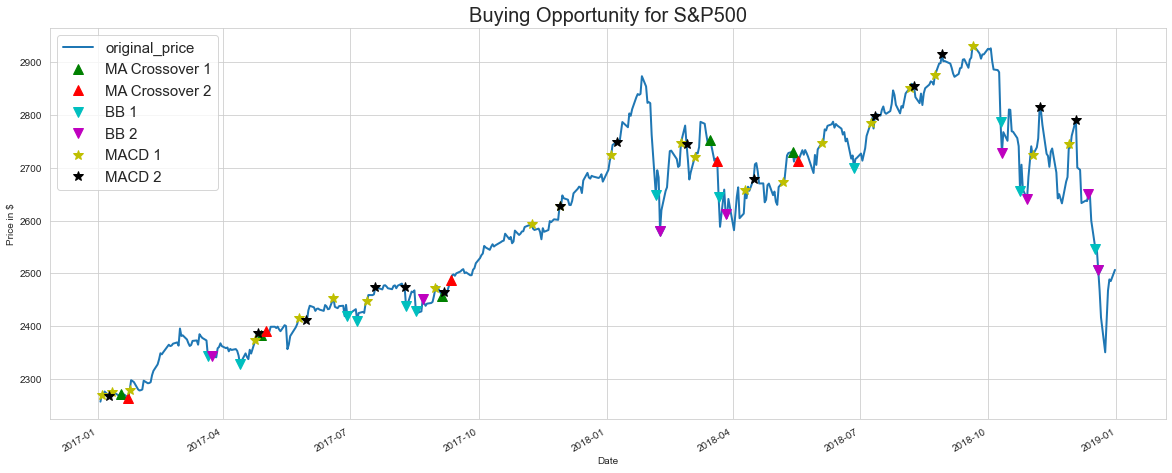}
\caption{The comparison of the buying opportunities generated by MA crossover, MACD and BB between workflow 1 and 2}\label{fig:trading}
\end{figure*}

\begin{table}
\footnotesize
\caption{Close Price Comparison via MA Crossover}
  \centering
\label{tab:ma}
\begin{tabular}{cc|cc}
\toprule
\textbf{Date} & \textbf{Buy with} & \textbf{Date} & \textbf{Buy with}\\ 
 &\textbf{Original Signals} & & \textbf{Denoised Signals}\\
\midrule
2017-01-18 & 2272 & 2017-01-23 & {\color[HTML]{FE0000} 2265} \\
2017-04-28 & 2384 & 2017-05-02 & 2391 \\
2017-09-05 & 2458 & 2017-09-11 & 2488  \\
2018-03-16 & 2752 & 2018-03-21 & {\color[HTML]{FE0000} 2712} \\
2018-05-14 & 2730 & 2018-05-18 & {\color[HTML]{FE0000} 2713} \\ 
\bottomrule
\end{tabular}
\end{table}

\begin{table}
\footnotesize
\caption{Close Price Comparison via BB}
  \centering
\label{tab:bb}
\begin{tabular}{cc|cc}
\toprule
\textbf{Date} & \textbf{Buy with} & \textbf{Date} & \textbf{Buy with}\\ 
 &\textbf{Original Signals} & & \textbf{Denoised Signals}\\
\midrule
2017-03-21  &  2344.02 & 2017-03-24  &  {\color[HTML]{FE0000}2343.98}\\
2017-04-13  &  2329 & & \\
2017-06-29  &  2420 & & \\
2017-07-06  &  2410 & &\\
2017-08-10  &  2438 & &\\
2017-08-17  &  2430 & 2017-08-22 & 2453\\
2018-02-05  &  2649 & & \\
2018-02-08  &  2581 & 2018-02-08  &  {\color[HTML]{009901}2581}\\
2018-03-22  &  2644 & 2018-03-27 & {\color[HTML]{FE0000}2613}\\
2018-06-27  &  2700 & & \\
2018-10-10  &  2786 & 2018-10-11 &  {\color[HTML]{FE0000}2728}\\
2018-10-24  &  2656 & 2018-10-29 &   {\color[HTML]{FE0000}2641}\\
2018-12-17  &  2546 & 2018-12-12 &   2651\\
2018-12-19  &  2507 & 2018-12-19  &  {\color[HTML]{009901}2507}\\ 
\bottomrule
\end{tabular}
\end{table}

\begin{table}
\footnotesize
\caption{Close Price Comparison via MACD}
  \centering
\label{tab:macd}
\begin{tabular}{cc|cc}
\toprule
\textbf{Date} & \textbf{Buy with} & \textbf{Date} & \textbf{Buy with}\\ 
 &\textbf{Original Signals} & & \textbf{Denoised Signals}\\
\midrule
2017-01-04  &  2271 & & \\
2017-01-11  &  2275 & 2017-01-09 & {\color[HTML]{FE0000}2267}\\
2017-01-24  &  2280 & & \\
2017-04-24  &  2374 & 2017-04-26 & 2387\\
2017-05-25  &  2415 & 2017-05-30 & {\color[HTML]{FE0000}2413}\\
2017-06-19  &  2453 & & \\
2017-07-13  &  2448 & 2017-07-19 & 2474\\
2017-08-31  &  2472 & 2017-08-09 & {\color[HTML]{3166FF}2474}\\
2017-11-08  &  2594 & & \\
2017-11-28  &  2627 & 2017-11-28  &  {\color[HTML]{009901}2627}\\
2018-01-04  &  2724 & 2018-01-08 & 2748\\
2018-02-23  &  2747 & 2018-02-27 & {\color[HTML]{FE0000}2744}\\
2018-03-05  &  2721 & & \\
2018-04-10  &  2657 & 2018-04-16 & 2678\\
2018-05-07  &  2673 & & \\
2018-06-04  &  2747 & & \\
2018-07-09  &  2784 & 2018-07-12 & 2798\\
2018-08-06  &  2850 & 2018-08-09 & {\color[HTML]{3166FF}2854}\\
2018-08-24  &  2875 & 2018-08-29 & 2914\\
2018-09-20  &  2931 & & \\
2018-11-02  &  2723 & 2018-11-07 & 2814\\
2018-11-28  &  2744 & 2018-12-03 & 2790\\
\bottomrule
\end{tabular}
\end{table}

\smallskip \noindent{\bf Trading opportunities and costs.}
Figure \ref{fig:trading} demonstrates the examples of applying technical indicators on S\&P 500 daily close price, where the buying opportunities are marked as green and red triangles, blue and purple reversed triangles, yellow and black stars for workflow 1 and 2 by using MA crossover, BB and MACD correspondingly. For the MA crossover indicator, the shorter simple moving average has a 10-day lookback window and the longer one has a 50-day lookback window. The lookback windows for both moving averages are chosen due to the small range of the dataset. It is clear in Table \ref{tab:ma} that the indicator triggers 5 buying opportunities based on the original close price and denoised close price, and there are three buying signals generated from the denoised close price having lower positions (highlighted in red colour). 
For the MACD indicator, there are 14 out of 22 buying signals produced based on the denoised price. It can be seen from Table \ref{tab:macd} that there are 3 buying signals (highlighted in red colour) from the denoised close price which propose lower buying positions, 1 buying signal (highlighted in green colour) proposes the same buying position and another 2 buying signals (highlighted in blue colour) propose positions that are higher but within 5 dollars. Though this indicator does not perform as well as we expected, there are still approximately a half of buying signals from the denoised close price proposed within a reasonable range. 
For the BB indicator, there are 8 buying signals generated based on the denoised close price among 14 signals based on the original close price. It can be seen from Table \ref{tab:bb} that 4 buying signals (highlighted in red colour) from the denoised close price propose lower buying positions and 2 of them (highlighted in green colour) suggest buying at the same position with the generated from the original close price. We thus can conclude that the majority of buying signals are correctly created by using the reconstructed price, and more than half of them suggest buying at lower prices. Moreover, the important buying opportunities are not accidentally smoothed out with noise.

\subsection{Large high-frequency dataset}
\begin{figure*}[ht]
  \centering
  \includegraphics[width=0.8\linewidth]{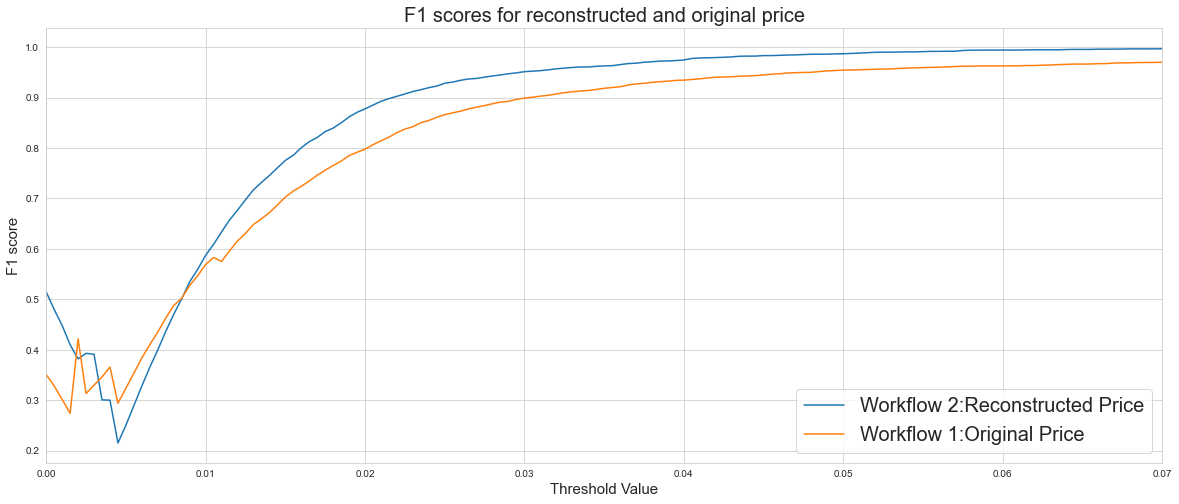}
  \caption{The F1 Scores Between Reconstructed and Original Price of 1-minute Bitcoin}\label{fig:btc}
\end{figure*}

\begin{figure*}[ht]
  \centering
  \includegraphics[width=0.8\linewidth]{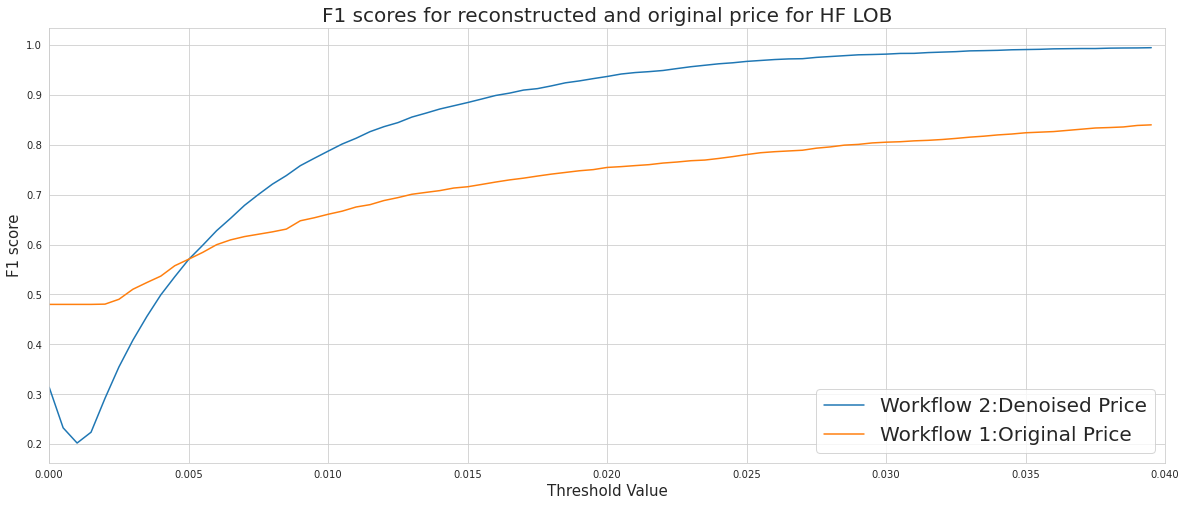}
  \caption{The F1 Scores Between Reconstructed and Original Price of 1-minute Bitcoin}\label{fig:hf}
\end{figure*}

We also examine the performance of the self-supervised learning model where the dataset is large and noisy. The same contrastive experiment is performed on 1-minute Bitcoin prices within a year and high-frequency limit order data of Google with a day. The comparison plot between the original price curve and the reconstructed price curve is hard to check visually due to the scale, but the F1 scores in Figure \ref{fig:btc} and Figure \ref{fig:hf} clearly reveal that self-supervised learning outperforms naive labeling with large and noisy datasets.

\section{Conclusions}
In this paper, we are motivated by the success in image recognition and deep neural networks from recent studies and we show that these techniques can be used for financial time-series classification problems. We also verify that the reconstructed price is meaningful since it can improve the accuracy of classification problems in financial labelling. The contrastive experiments are performed on both small and large noisy datasets, and we demonstrate the advantages of using self-supervised learning in identifying the trading opportunities on S\&P500 daily close price. Our study suggests that self-supervised learning can outperform the traditional models on real financial time-series, and the denoised close price can be adopted to create new strategies for backtesting.

Several directions for future research exist. It remains to compare the trading signals triggered by the original 1-minute Bitcoin close price and high-frequency limit order data and their denoised prices. The challenge of designing a trading algorithm that performs well on the denoised prices and can be backtested in a variety of conditions is also an interesting problem. 
%
More generally, the model could be further improved. Possible directions include: 1) replace the \emph{naive} labelling method with volatility-based approaches, such as the triple-barrier method \cite{machine_learning_book}; 2) transfer the financial time-series into 2-dimensional data to use 2-d convolutional neural network for denoising autoencoder \cite{9080613}; 3) examine the model's performance with other pretext tasks or downstream tasks, for example, using stacked autoencoder as the pretext task while remaining the downstream task unchanged. It is also worth exploring different combinations between pretext tasks and downstream tasks.

\bibliographystyle{unsrt}  
\bibliography{sample-base}

\begin{thebibliography}{10}

\bibitem{image1}
Naftali Cohen, Tucker Balch, and Manuela Veloso.
\newblock Trading via image classification.
\newblock {\em CoRR}, abs/1907.10046, 2019.

\bibitem{self1}
Longlong Jing and Yingli Tian.
\newblock Self-supervised visual feature learning with deep neural networks:
  {A} survey.
\newblock {\em CoRR}, abs/1902.06162, 2019.
\newblock \url{http://arxiv.org/abs/1902.06162}.

\bibitem{murphy1999technical}
John~J. Murphy.
\newblock {\em Technical analysis of the financial markets}.
\newblock New York Institute of Finance, New York, NY, 1999.

\bibitem{noise}
Fischer Black.
\newblock Noise.
\newblock {\em The Journal of Finance}, 41(3):528--543, July 1986.

\bibitem{DBLP:journals/corr/ShelhamerLD16}
Evan Shelhamer, Jonathan Long, and Trevor Darrell.
\newblock Fully convolutional networks for semantic segmentation.
\newblock {\em CoRR}, abs/1605.06211, 2016.

\bibitem{DBLP:journals/corr/abs-1801-00868}
Alexander Kirillov, Kaiming He, Ross~B. Girshick, Carsten Rother, and Piotr
  Doll{\'{a}}r.
\newblock Panoptic segmentation.
\newblock {\em CoRR}, abs/1801.00868, 2018.

\bibitem{machine_learning_book}
Marcos~Lopez de~Prado.
\newblock {\em Advances in Financial Machine Learning}.
\newblock John Wiley \& Sons, New York, NY, USA, February 2018.

\bibitem{DBLP:journals/corr/abs-1810-04805}
Jacob Devlin, Ming{-}Wei Chang, Kenton Lee, and Kristina Toutanova.
\newblock {BERT:} pre-training of deep bidirectional transformers for language
  understanding.
\newblock {\em CoRR}, abs/1810.04805, 2018.

\bibitem{DBLP:journals/corr/abs-1811-06964}
Eric Jang, Coline Devin, Vincent Vanhoucke, and Sergey Levine.
\newblock Grasp2vec: Learning object representations from self-supervised
  grasping.
\newblock {\em CoRR}, abs/1811.06964, 2018.

\bibitem{DBLP:journals/corr/abs-1901-09005}
Alexander Kolesnikov, Xiaohua Zhai, and Lucas Beyer.
\newblock Revisiting self-supervised visual representation learning.
\newblock {\em CoRR}, abs/1901.09005, 2019.

\bibitem{DBLP:journals/corr/abs-1804-03641}
Andrew Owens and Alexei~A. Efros.
\newblock Audio-visual scene analysis with self-supervised multisensory
  features.
\newblock {\em CoRR}, abs/1804.03641, 2018.

\bibitem{denosing_example}
Pascal Vincent, Hugo Larochelle, Yoshua Bengio, and Pierre-Antoine Manzagol.
\newblock Extracting and composing robust features with denoising autoencoders.
\newblock In {\em In Proceedings of the 25th international conference on
  Machine learning}, page 1096–1103, 2018.

\bibitem{DBLP:journals/corr/ZhangIE16}
Richard Zhang, Phillip Isola, and Alexei~A. Efros.
\newblock Colorful image colorization.
\newblock {\em CoRR}, abs/1603.08511, 2016.

\bibitem{DBLP:journals/corr/PathakKDDE16}
Deepak Pathak, Philipp Kr{\"{a}}henb{\"{u}}hl, Jeff Donahue, Trevor Darrell,
  and Alexei~A. Efros.
\newblock Context encoders: Feature learning by inpainting.
\newblock {\em CoRR}, abs/1604.07379, 2016.

\bibitem{DBLP:journals/corr/abs-1807-03748}
A{\"{a}}ron van~den Oord, Yazhe Li, and Oriol Vinyals.
\newblock Representation learning with contrastive predictive coding.
\newblock {\em CoRR}, abs/1807.03748, 2018.

\bibitem{doi:10.1142/S0218194016400088}
Daoyuan Li, Tegawende~F. Bissyande, Jacques Klein, and Yves~Le Traon.
\newblock Time series classification with discrete wavelet transformed data.
\newblock {\em International Journal of Software Engineering and Knowledge
  Engineering}, 26(09n10):1361--1377, 2016.

\bibitem{10.1007/978-3-540-24775-3_71}
Hui Zhang, Tu~Bao Ho, and Mao~Song Lin.
\newblock A non-parametric wavelet feature extractor for time series
  classification.
\newblock In Honghua Dai, Ramakrishnan Srikant, and Chengqi Zhang, editors,
  {\em Advances in Knowledge Discovery and Data Mining}, pages 595--603,
  Berlin, Heidelberg, 2004. Springer Berlin Heidelberg.

\bibitem{StatisticalMethods}
Daniel~S. Wilks.
\newblock {\em Statistical Methods in the Atmospheric Sciences}, volume 100.
\newblock Elsevier Science \& Technology, 2011.

\bibitem{cite-key}
Anthony Bagnall, Jason Lines, Aaron Bostrom, James Large, and Eamonn Keogh.
\newblock The great time series classification bake off: a review and
  experimental evaluation of recent algorithmic advances.
\newblock {\em Data Mining and Knowledge Discovery}, 31(3):606--660, 2017.

\bibitem{deeplearningreview}
Hassan~Ismail Fawaz, Germain Forestier, Jonathan Weber, Lhassane Idoumghar, and
  Pierre{-}Alain Muller.
\newblock Deep learning for time series classification: a review.
\newblock {\em CoRR}, abs/1809.04356, 2018.

\bibitem{doi:10.1080/14697688.2019.1622295}
Justin Sirignano and Rama Cont.
\newblock Universal features of price formation in financial markets:
  perspectives from deep learning.
\newblock {\em Quantitative Finance}, 19(9):1449--1459, 2019.

\bibitem{8081663}
Avraam Tsantekidis, Nikolaos Passalis, Anastasios Tefas, Juho Kanniainen,
  Moncef Gabbouj, and Alexandros Iosifidis.
\newblock Using deep learning to detect price change indications in financial
  markets.
\newblock In {\em 2017 25th European Signal Processing Conference (EUSIPCO)},
  pages 2511--2515, 2017.

\bibitem{10.1371/journal.pone.0180944}
Wei Bao, Jun Yue, and Yulei Rao.
\newblock A deep learning framework for financial time series using stacked
  autoencoders and long-short term memory.
\newblock {\em PLOS ONE}, 12(7):1--24, 07 2017.

\bibitem{Park_2019}
Daniel~S. Park, William Chan, Yu~Zhang, Chung-Cheng Chiu, Barret Zoph, Ekin~D.
  Cubuk, and Quoc~V. Le.
\newblock Specaugment: A simple data augmentation method for automatic speech
  recognition.
\newblock {\em Interspeech 2019}, Sep 2019.

\bibitem{Wang2014EncodingTS}
Zhiguang Wang and T.~Oates.
\newblock Encoding time series as images for visual inspection and
  classification using tiled convolutional neural networks.
\newblock In {\em Association for the Advancement of Artificial Intelligence},
  2014.

\bibitem{DBLP:journals/corr/WangO15}
Zhiguang Wang and Tim Oates.
\newblock Imaging time-series to improve classification and imputation.
\newblock {\em CoRR}, abs/1506.00327, 2015.

\bibitem{DBLP:journals/corr/abs-1901-05237}
Yun{-}Cheng Tsai, Jun~Hao Chen, and Chun{-}Chieh Wang.
\newblock Encoding candlesticks as images for patterns classification using
  convolutional neural networks.
\newblock {\em CoRR}, abs/1901.05237, 2019.

\bibitem{9222178}
Dan Wang, Tianrui Wang, and Ionuţ Florescu.
\newblock Is image encoding beneficial for deep learning in finance?
\newblock {\em IEEE Internet of Things Journal}, pages 1--1, 2020.

\bibitem{DBLP:journals/corr/abs-2002-09545}
Jingkun Gao, Xiaomin Song, Qingsong Wen, Pichao Wang, Liang Sun, and Huan Xu.
\newblock Robusttad: Robust time series anomaly detection via decomposition and
  convolutional neural networks.
\newblock {\em CoRR}, abs/2002.09545, 2020.

\bibitem{bitcoin}
Bitcoin btc.
\newblock \url{https://www.coindesk.com/price/bitcoin}, Accessed Jun. 16, 2021.

\bibitem{hflob}
Lobster academic data.
\newblock \url{https://lobsterdata.com/info/DataSamples.php }, Accessed Aug.
  16, 2021.

\bibitem{Goodfellow-et-al-2016}
Ian Goodfellow, Yoshua Bengio, and Aaron Courville.
\newblock {\em Deep Learning}.
\newblock MIT Press, 2016.
\newblock Chapter 14. \url{http://www.deeplearningbook.org}.

\bibitem{Encyclopedia}
Robert~W. Colby and Thomas~A. Meyers.
\newblock {\em The encyclopedia of technical market indicators}.
\newblock Homewood, Ill: Dow Jones-Irwin, 1988.

\bibitem{9080613}
Silvio Barra, Salvatore~Mario Carta, Andrea Corriga, Alessandro~Sebastian
  Podda, and Diego~Reforgiato Recupero.
\newblock Deep learning and time series-to-image encoding for financial
  forecasting.
\newblock {\em IEEE/CAA Journal of Automatica Sinica}, 7(3):683--692, 2020.

\end{thebibliography}

\end{document}